%% file: main.tex
\documentclass[final]{cvpr}

\usepackage{times}
\usepackage{epsfig}
\usepackage{graphicx}
\usepackage{amsmath}
\usepackage{amssymb}
\usepackage{kotex}
\usepackage[dvipsnames]{xcolor}
\usepackage{multirow}
\usepackage{makecell}
\newif\ifdraft\draftfalse

\ifdraft

\newcommand\hj[1]{\textcolor{DarkOrchid}{#1}}
\newcommand\jy[1]{\textcolor{Sepia}{#1}}
\newcommand\sh[1]{\textcolor{ForestGreen}{#1}}
\newcommand\nj[1]{\textcolor{red}{#1}}

\else

\newcommand\hj[1]{#1}
\newcommand\jy[1]{#1}
\newcommand\sh[1]{#1}
\newcommand\nj[1]{#1}

\fi
  {\begin{list}{}%
          {\setlength{\leftmargin}{#1}}%
          \item[]%
  }
  {\end{list}}
  
\usepackage[pagebackref=true,breaklinks=true,colorlinks,bookmarks=false]{hyperref}

\def\cvprPaperID{****} % *** Enter the CVPR Paper ID here
\def\confYear{CVPR 2021}

\begin{document}

%%%%%%%%% TITLE
\title{Learning Dynamic Network Using a Reuse Gate Function in \\ Semi-supervised Video Object Segmentation}
%Does not Need to Fully-calculate Every Frame for \\  Semi-supervised Video Object Segmentation}

% Learning when to skip layers using a reuse gate function in Semi-supervised Video Object Segmentation
% Learning dynamic network using a reuse gate function in Semi-supervised Video Object Segmentation
% Training dynamic network using a reuse gate function in Semi-supervised Video Object Segmentation -> (1)
% Dynamic inference using a reuse gate function in Semi-supervised Video Object Segmentation

\author{
	\small
	\begin{tabular}{c c c c c}                              
	    \bf Hyojin Park$^1$ &
	    \bf Jayeon Yoo$^1$ \thanks{Indicate  equal contribution as second authors}&
	    \bf Seohyeong Jeong$^{1,3}$ \footnotemark[1] \thanks{This work was done when Seohyeong Jeong was with SNU}&
	    \bf Ganesh Venkatesh$^2$ &
	    \bf Nojun Kwak$^1$ \\
	                              
		\multicolumn{5}{c}{$^1$Seoul National University, $^2$Facebook Inc.,  $^3$AIRS Company, Hyundai Motor Group } \\                                              

	\end{tabular}                                                                       
} 
\maketitle

%%%%%%%%% ABSTRACT
\begin{abstract}
Current state-of-the-art approaches for Semi-supervised Video Object Segmentation (Semi-VOS) propagates information from previous frames to generate segmentation mask for the current frame. 
This results in high-quality segmentation across challenging scenarios such as changes in appearance and occlusion.
But it also leads to unnecessary computations for stationary or slow-moving objects where the change across frames is minimal.
In this work, we exploit this observation by using temporal information to quickly identify frames with minimal change and skip the heavyweight mask generation step.
To realize this efficiency, we propose a novel dynamic network that estimates change across frames and decides which path -- computing a full network or reusing previous frame's feature -- to choose depending on the expected similarity.
Experimental results show that our approach significantly improves inference speed without much accuracy degradation on challenging Semi-VOS datasets -- DAVIS 16, DAVIS 17, and YouTube-VOS. 
Furthermore, our approach can be applied to multiple Semi-VOS methods demonstrating its generality. 
The code is available in https://github.com/HYOJINPARK/Reuse\_VOS .

\end{abstract}

%%%%%%%%% BODY TEXT
\input{Texs/Introv2}
\input{Texs/Relate_sh}

\input{Texs/Method1}
\input{Texs/Method2_v2}

\input{Texs/Method3}

\input{Texs/Method4}
\input{Texs/Exp1}

\input{Texs/Exp2}
\input{Texs/Conc_sh}
\section*{ACKNOWLEDGMENTS}
\label{ack}
This work was supported by the National Research Foundation of Korea (NRF) grant funded by the Korea government (2021R1A2C3006659).

{\small
\bibliographystyle{ieee_fullname}
\bibliography{egbib}
}

\end{document}

%% file: Texs/Introv2.tex
\section{Introduction}
\label{sec:intro}

Semi-VOS tracks an object of interest across all the frames in a video given the ground truth mask of the initial frame. 
VOS classifies each pixel as belonging to background or a tracked object. This task has wide applicability to many real-world use cases including autonomous driving, surveillance, video editing as well as to the emerging class of augmented reality/mixed reality devices.
VOS is a challenging task because it needs to distinguish the target object from other similar objects in the scene even as target's appearance changes over time as well as through occlusions.

\begin{figure}[t]
\begin{center}
\includegraphics[width=0.9\linewidth]{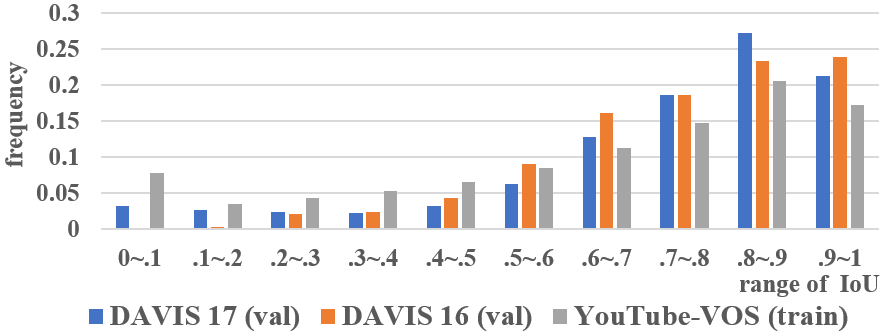}\\
(a)
\\
\includegraphics[width=0.9\linewidth]{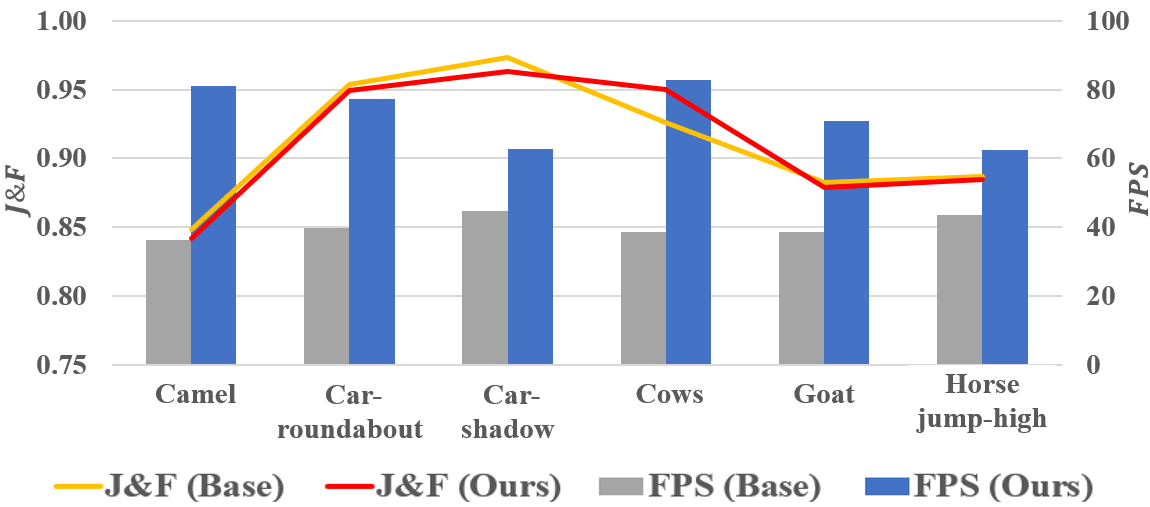}\\
(b) 
\\
\end{center}
   \caption{(a) Histogram of a range of IoU between the previous and the current ground truth masks. 
   X-axis is a range of IoU and y-axis denotes frequency corresponding to the range of IoU.
  (b) FPS and accuracy ($J\&F$) comparison between the baseline model (FRTM \cite{robinson2020learning}) and our dynamic network on videos having high IoU between the previous and current ground truth masks.
  The proposed method preserves the original accuracy while improving the speed a lot.
   }
\label{fig:teaserIntro2}
\end{figure}

A variety of methods have been proposed for solving video object segmentation including online learning~\cite{robinson2020learning, maninis2018video}, mask propagation~\cite{lin2020flow, cheng2017segflow}, and template matching~\cite{wang2019fast, oh2019video}. 
A common theme across most of these previous methods is to use information from previous frames -- either just the first frame, some of the previous frames (first and last being a popular option) or all the previous frames -- to produce high quality segmentation mask. 

In this work, we ask a different question\\ \textbf{Q: Can we use temporal information to identify when the object appearance and position has not changed across frames?}\\
The motivation for doing so would be to skip much of the expensive computation needed to produce a high-quality mask for the current frame if that mask is almost the same as the mask we computed in the previous frame.
Instead, we can produce current frame's mask using a cheap model that just makes minor edits to the previous frame's feature.
As shown in Fig.~\ref{fig:teaserIntro2}(a), for a significant fraction of the frames in the popular DAVIS and YouTube-VOS dataset, object masks are very similar to their previous frame's masks (73.3\% of consecutive frames in DAVIS 17 dataset have IoU greater than 0.7).

We build on the above observation by constructing a cheap temporal matching module to quickly quantify the similarity of the current frame with the previous frame.
We use the similarity to gate the computation of high-quality mask for the current frame -- if the similarity is high we reuse previous features with minor refinements and avoid the expensive mask generation step.
This allows us to avoid majority of computations for the current frame without compromising on accuracy.

Our approach compliments the existing video object segmentation approaches and to demonstrate its generality, we integrate our proposal into multiple prior video object segmentation models -- FRTM~\cite{robinson2020learning} and TTVOS~\cite{park2020ttvos}. To the best of our knowledge, we are the first to propose skipping computation of segmentation masks dynamically based on the object movement. We believe this is a significant contribution that will enable high-quality video object segmentation models to run on mobile devices in real time with minimal battery impact.

We make the following contributions in this paper:
\begin{itemize}
    \item We make the case for exploiting temporal information to skip mask generation for frames with little or no movement.
    \item We develop a general framework to skip mask computation consisting of sub-networks to estimate movement across frames, dynamic selection between processing full-network or reusing previous frame's feature for generating mask and a novel loss function to train this dynamic architecture.
    \item We evaluate our approach on multiple video object segmentation models (FRTM, TTVOS) as well as multiple challenging datasets (DAVIS 16, DAVIS 17, Youtube-VOS) and demonstrate that we can save up to $47.5\%$ computation and speedup FPS by $1.45\times$ with minimal accuracy impact on DAVIS 16 (within around 0.4 $\%$ of baseline). 
    
    % \item \hj{We develop a general framework to skip sub-network by exploiting temporal information for identifying little movement (template matching module).
    % \item Our framework selects between different paths (full mask generation vs adjustment over previous features) by a reuse gate function\footnote{We use the term a ``reuse gate function" to avoid ambiguity. In previous works, a gate function allows model to skip layers when the gate is off. However, in our work, a model skips layers when the gate is on.}
    % \item We propose \sh{a} novel loss functions to train the template matching module and the gate function.
    % \item We evaluate our approach on multiple video object segmentation models (FRTM, TTVOS) as well as multiple challenging datasets (DAVIS16, DAVIS17, YouTube-VOS) and demonstrate that we achieves speed-up FPS with minimal accuracy impact. }
    % \nj{\# Check the numbers!}
\end{itemize}

%% file: Texs/Relate_sh.tex
\section{Related Work}
\label{Relate}

\noindent
\textbf{Online-learning: }
Online-learning algorithms learn to update models from datastreams in sequential manners during the inference stage~\cite{ijcai2018-369,zhou2012online,kivinen2004online}.
In the semi-VOS task, online learning takes place as fine-tuning the segmentation model during the inference stage given the image and the target mask of the first frame to inject the strong appearance of the mask to the model~\cite{maninis2018video,perazzi2017learning,Cae+17}. 
However, the fine-tuning step causes a significant bottleneck.
FRTM~\cite{robinson2020learning} tackles this issue by splitting the model into two sub-networks: a light-weight target appearance model trained online and a segmentation network trained offline. 

\noindent
\textbf{Mask Propagation: }
Mask propagation methods realigns the given segmentation mask or features. % using the estimated flow vector. 
Optical flow is widely used to measure the changes in pixel-wise movements of objects in VOS~\cite{khoreva2017lucid,dutt2017fusionseg,tsai2016video,sevilla2016optical}. Segflow \cite{cheng2017segflow} designs two branches of image segmentation and optical flow, and bidirectionally combines both information into a unified framework to estimate target masks. Similarly, FAVOS \cite{lin2020flow} and CRN \cite{hu2018motion} utilize optical flow information to refine a coarse segmentation mask into an accurate mask.

\noindent
\textbf{Template matching: }
Template matching is one of the common approaches in the semi-VOS domain.
In template matching, models generate a target template and calculate similarity between the template and given inputs.
A-GAME \cite{johnander2019generative} employs a mixture of Gaussians to learn the target and background feature distributions. RANet \cite{wang2019ranet} integrates a ranking system to the template matching process to rank and select feature maps according to their importance for fine-grained VOS performance.
FEELVOS \cite{voigtlaender2019feelvos} calculates a distance map by local and global matching mechanism to transfer previous information to the current frame for better robustness.
Furthermore, SiamMask \cite{wang2019fast} exploits a depth-wise operation to make the matching operation faster.
% The memory networks approach is widely used in question answering tasks for handling long-term sequential data~\cite{kim2019progressive,sukhbaatar2015end,weston2014memory}. % Q: memory network has to do with template matching?
% STM \cite{oh2019video} and  GC \cite{li2020fast} integrate the idea for semi-VOS task. % Q: are they QA models??
STM \cite{oh2019video} and  GC \cite{li2020fast} integrate the memory network approach~\cite{kim2019progressive,sukhbaatar2015end,weston2014memory}.
However, this approach requires lots of resources in maintaining the memory. 
TTVOS~\cite{park2020ttvos} proposes a light-weight template matching method for reducing the burden of computation where the temporal consistency loss is used to endow a correction power about the incorrectly estimated mask to the model.
They claim that measuring the exact optical flow for temporal consistency is too demanding, 
and they estimate a transition matrix to identify the regions changing from background to foreground and vice versa.

\begin{figure*}[t]

\begin{center}
    \includegraphics[width=0.90\linewidth]{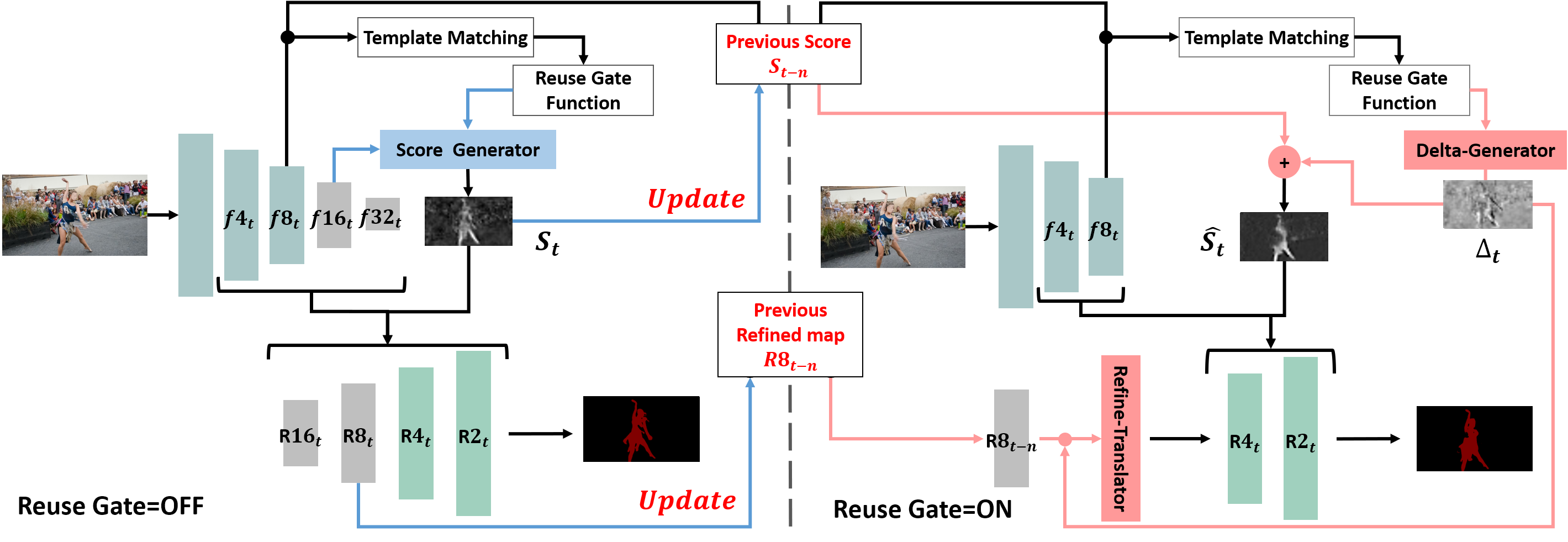}
\end{center}
   \caption{Overall architecture of our dynamic model.
   Some parts of feature extraction and segmentation network are skipped when the reuse gate is on.
   The delta-generator produces $\Delta_t$ to convert the previous feature information to that of the current feature.
   Refine-translator transforms the previous refined feature map into the current one with the help of $\Delta_t$ to make the final mask.
   When the reuse gate is off, the previous score map is updated to be used in next frames.}
\label{fig:arch}
\end{figure*}

\noindent
\textbf{Dynamic network: }
Dynamic networks perform efficient inference by dynamically choosing a subset of networks depending on the input. 
Constructing a dynamic inference path by dropping sub-layers of a network using gating modules has been widely studied in image-level tasks~\cite{Zuxuan_2018_CVPR, Ravi_2018_CVPR,veit2018convolutional,lee2020urnet}. This approach has been applied to image segmentation~\cite{Yanwei_2020_CVPR} and has been recently extended to the video domain as well~\cite{chen2020state,hu2020temporally, Zhang_2019_ICCV, li2018low, shelhamer2016clockwork}.
% In video-level tasks, especially in a segmentation problem, processing the entire frames of a video demands heavy computation.
% However, \cite{chen2020state, Zhang_2019_ICCV} is just switching modules and still requires the full computation on the feature extraction and the mask refinement for every frame.
\cite{chen2020state, Zhang_2019_ICCV} adopt switching modules to Semi-VOS field, but they still requires the full computation on the feature extraction and the mask refinement for every frame.

We find two problems about applying dynamic network approaches to Semi-VOS field.
First, the gate function outputs a discrete decision which are hard to be integrated with a convolutional network due to the difficulty of gradient calculation.
Gumbel-Softmax trick \cite{jang2016categorical} is generally used to resolve the problem by the softmax relaxation, but we empirically get unstable training problem.
The second issue is how to penalize the model.
Usually, there is a target rate for training a dynamic network.
The target rate determines the fixed maximum number of layer blocks that can be used in the computation during training.
We found that this constraint does not work well in this task.
Since the target rate drives a model to concentrate on meeting the desired number of gates on, the model tends to treat preserving the original accuracy as a less important matter, resulting in a poor segmentation accuracy.

%% file: Texs/Method1.tex
\section{Method}
\label{method}

We explain our dynamic architecture that estimates the movement across frames and skip mask generation for the VOS task.
In Sec \ref{sec:arch}, we briefly summarize the baseline model, FRTM\cite{robinson2020learning}, and introduce our method of converting the baseline model into a dynamic architecture.
Note that our proposed dynamic inference architecture can be applied to other VOS frameworks as well.
In Sec \ref{sec:template}, we explain our template matching method for measuring dissimilarity. The reuse gate function takes the dissimilarity information to quantify movement across frames and selects between different paths for generating mask (processing full-network
when the reuse gate is off and reusing previous frame's feature when the reuse gate is on). In the case of the reuse gate being on, the model produces a difference map between adjacent frames using the delta-generator.
In Sec \ref{sec:refine}, we show how the refine-translator adjusts the previous refined feature map, $R_{t-n}$, to the current refined feature map, $\hat{R}_{t}$.
Finally, in Sec \ref{sec:gating}, we have empirically witnessed that when the common constraint for the gate function~\cite{lee2020urnet,veit2018convolutional} is used, the model experiences dramatic performance degradation.
To resolve this problem, we introduce a new gating loss, called gating probability loss, that takes the IoU between the current and the previous masks into account as described. 

\subsection{Overall Architecture}
\label{sec:arch}
\noindent
\textbf{Previous work: }
Online learning methods train models to learn target-specific appearance during the inference stage.
This enhances the robustness of a model but it still suffers from extensive latency due to the fine-tuning step.
FRTM \cite{robinson2020learning} depicted in Fig. \ref{fig:arch} (excluding the proposed \textit{template matching} and \textit{gate function} modules) resolves this chronic issue in the online-learning realm by splitting the model into a light-weight score generator and a segmentation network.
The light-weight score generator simply consists of two layers for faster optimization during online learning, and it produces a coarse score map of an object.
The segmentation network is much more complex than the light-weight module. 
Taking extracted feature maps as an extra input along with the score map, the network refines the coarse score map and generates high quality masks.
The segmentation network is trained offline to reduce the burden of online-learning.

As shown in Fig.~\ref{fig:arch}, a shared feature extractor produces feature maps $fN_t$ from the current frame, where $fN_t$ denotes a feature map at frame $t$ with an $1/N$-sized width and height of the input.
$f16_t$ is forwarded to the light-weight score generator to generate a target score map.
Then, the segmentation network gradually increases the spatial size of the feature map from $f32_t$ using the score map in a U-Net-like structure.
${f32_t, f16_t,f8_t}$ and $f4_t$ are enhanced for generating a more accurate mask by the score map.
The final high resolution feature map is converted to a target segmentation mask.
The second layer in the score generator is updated every eight frames to handle the changing of the target appearance.

\noindent
\textbf{Our work: }
We analyze that a substantial amount of frames in a video are similar to each others and for these redundant frames, the model can reuse previous information instead of full path calculation.
Therefore, in our model, not every layer need to be fully-forwarded to extract and refine features.
The template matching is applied to measure movement from dissimilarity between current and previous frames.
The gate function decides whether to skip calculation or not from an output feature map of the template matching module.
% the embedding misalignment between the current feature map $f8_t$ and the previous score map $S_{t-n}$.
Details are described in Sec \ref{sec:template}.
% Here, the score map contains information of blahblah.}
% The feature map and previous score map are concatenated to enhance the feature map by following the score map about the foreground region. % ?
%%%%%%%%%%%%%%%%%%%%%%%%%%%%%%%%%%%%%%%%%%%%%%%%%%%%%%%%%%%%%%%%%%
% The feature map and previous score map are concatenated and passed on to the template matching module to attend on the foreground region of the score map.
% The greater the difference between the target and the previous frame, the more the dis-aligned information is included to the enhanced feature map.
% We impose the template matching stage to catch dis-alignment information between the current feature map and previous score map.
% An output feature map from the template matching is used as an input to a gate function to decide whether to skip calculation or not.

Our gate function consists of two convolution layers and two max-pooling layers as follows:
\begin{equation}
    \label{eq:gate}
     P_{gate} =\sigma (w_{2}*f(w_{1}*f(x))),
\end{equation}
where $w$ and $f$ denotes convolution weights and max-pooling in a layer, respectively and $\sigma$ is a sigmoid function that returns the probability, $P_{gate}$, of the gate being on (reuse).
If $P_{gate}$ is higher than a threshold $\tau$\footnote{In the training stage, we set $\tau = 0.5$.}, which means the current and previous frames are similar enough, the reuse gate is on as follows:
\begin{equation}
 \label{eq:switch}
g_t = \begin{cases}
1\; (\text{reuse})&\text{if $P_{gate} \ge \tau$}\\
0\; (\text{not reuse}) &\text{otherwise} %\text{if $P_{gate} < \tau$}.
\end{cases}
\end{equation}
%%%%%% gate=1 : p > tau , gate=0 : p < tau
%%%% when tau=0 => when estimated similarity is greater than 0 => every feature is reused
%%%% when tau=1 =>  when estimated similarity is greater than 1 => every feature is recalculated
% \hj{This means the current and previous frames are different enough.}
% If the gate is open, the current and previous frames are different enough. 
%\nj{구현과 상관 없이 $p$가 높으면 gate가 1, $p$가 낮으면 gate가 0이 된다고 기술하면 될 것 같음. 마찬가지로 $\tau$도 dissimilarity가 아닌 similarity로 바꾸었음. }

If the reuse gate is off ($g_t = 0$), the model generates a score map for the current frame following the original FRTM method and the generated score map is stored to be used as a previous score map for subsequent frames.
On the other hand, if the reuse gate is on, the model makes a delta map, $\Delta_t$, 
through the delta-generator.
The delta-map contains information on pixel-wise foreground-background conversion from the previous to the current frames.
The model adds the delta-map into the previous score to estimate the current score map.
Therefore, we can skip the remaining feature extraction process of calculating $f16_t$ and $f32_t$ which need to get score map in baseline.
In the segmentation network, we cannot use the original network due to missing $f16_t$ and $f32_t$ as shown in Fig. \ref{fig:arch}.
$RN_t$ denotes a refined feature map at frame $t$ with an $1/N$-sized width and height of the input, and is produced from $fN'_t$ and $S_t$, where $N'=N/2$.
We estimate $\hat{R8}_{t}$ by using the refine-translator
The previous refined feature map $R8_{t-n}$, and $\Delta_t$ are passed on to the refine-translator, and refine-translator consists of multiple size receptive fields to estimate $\hat{R8}_{t}$.
Therefore, the model can also skip stages of making $R16_t$ and $R8_t$. 
Details are described in Sec \ref{sec:refine}.

%% file: Texs/Method2_v2.tex
\subsection{Template Matching for Quantifying Movement} 
\begin{figure}[t]

\begin{center}
    \includegraphics[width=0.9\linewidth]{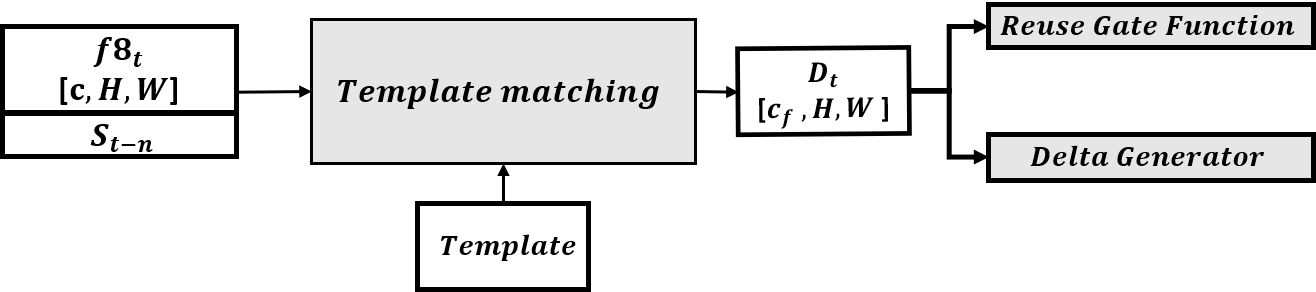}
\end{center}

  \caption{Process of the template matching. The output feature map of the template matching module, $D_t$, focuses on dissimilarities between the current and the previous frames. $D_t$ is forwarded to the gate function and the delta-generator.}
\label{fig:template}
\end{figure}

% \subsection{Template Matching for Misalignment Info.} 
\label{sec:template}
In order to quantify the movements across frames, we find dissimilarity information that is measured by utilizing a simple module introduced in TTVOS~\cite{park2020ttvos}.
They proposed a light-weight template matching module\sh{,} which generate\sh{s} a similarity map, as an output, to focus on the target from the input by comparing with a \textit{template}. The template contains the target appearance.
% %ing 
% \sh{a} similarity map and a transition matrix\sh{,} 
% % which 
% \sh{that} contains the information of the region changing from the background to the foreground and vice versa.
Inspired by this light-weight module, we integrate the template matching procedure into our framework to generate the dissimilarity feature, $D_t$ in Fig.~\ref{fig:template}, with some modifications. % \hj{for different purpose from TTVOS}.

%%%%%%%%%%%%%%%%%%%%%%%%%%%%%%%%%%%%%%%%%% Figure for template matching %%%%%%%%%%%%%%%%%%%%%%%%%%%%%%%%%%%%%%%%%%%%%%%%

\begin{figure}[t]

\begin{center}
    \includegraphics[width=0.9\linewidth]{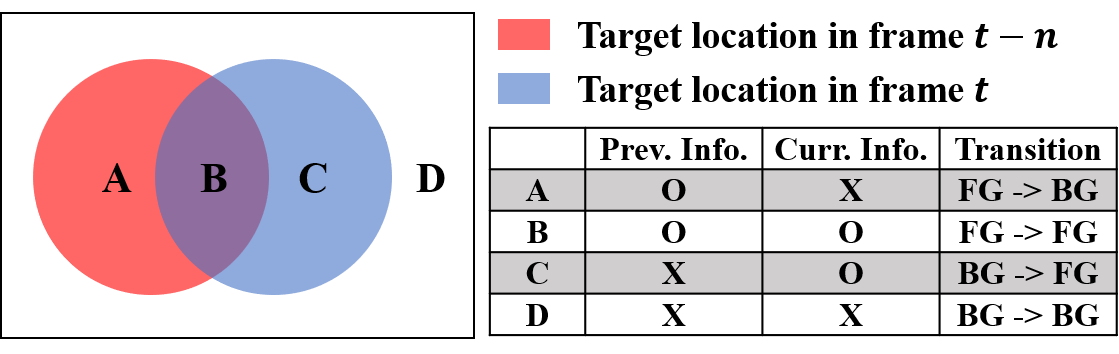}
\end{center}

   \caption{Example of template matching process. `Prev. info.' is represented by $S_{t-n}$. `Curr. info.' is represented by $f8_t$.
%   whether a region of the current image has target information or not. 
FG and BG denotes foreground and background, respectively.
   Transition indicates change in class from $t-n$ to $t$. 
   We focus to correctly identifying the dissimilar regions (A and C) between frames.}
\label{fig:diag}
\end{figure}

%%%%%%%%%%%%%%%%%%%%%%%%%%%%%%%%%%%%%%%%%% Figure for template matching %%%%%%%%%%%%%%%%%%%%%%%%%%%%%%%%%%%%%%%%%%%%%%%%

%%%%%%%%%%%%%%%%%%%%%%%%%%%%%%%%%% Fig Refine ######################################
\begin{figure*}[t]
\begin{center}
    \includegraphics[width=0.9\linewidth]{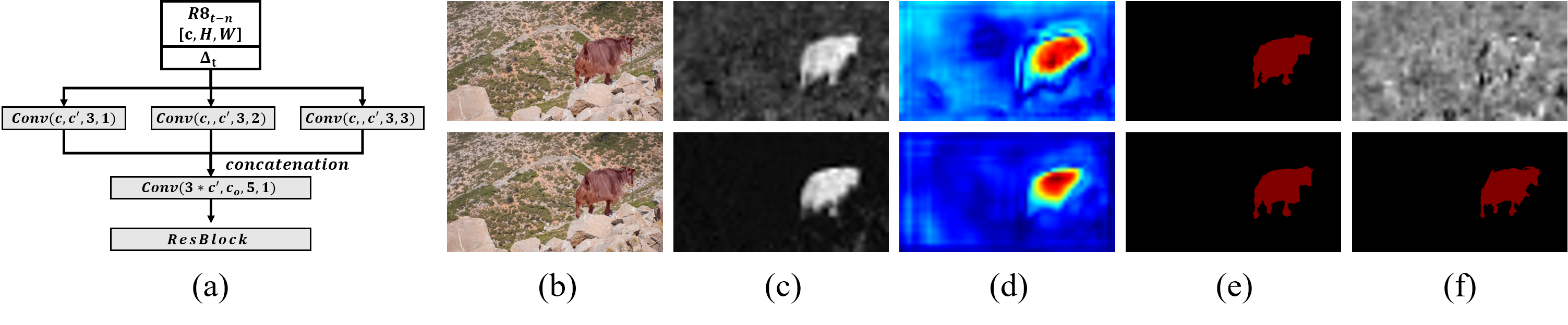} 
\end{center}

   \caption{(a) Refine-translator. $Conv(c,c',k,d)$ denotes a $d$ dilated $k\times k$ convolution layer with an input channel of $c$ and an output channel of $c'$. (b)-(e) Example of \textit{goat}. Top row is frame 24 and the bottom row is frame 25. (b) Input frames are overlapped with ground truth masks. (c) $S_{24}$ and $\hat{S}_{25}$. (d) Activation maps of $R8_{24}$ and $\hat{R8}_{25}$. (e) estimated masks. (f) Top: $\Delta_{25}$, Bottom: ground truth mask of frame $25$.}
\label{fig:RefineT}
\end{figure*}
%%%%%%%%%%%%%%%%%%%%%%%%%%%%%%%%%% Fig Refine ######################################

In our work, % instead of focusing on calculating similarity, 
we apply the property of template matching to focus on the dissimilarity, as described in Fig~\ref{fig:diag}.
% In the original template matching module, the gap between the calculated similarity map and the given ground truth mask decreases as the training process continues.
% However, our module finds dissimilarity for quantifying movement\sh{s}.
% To do this, we use A, 수식, and B, 수식..
To do this, we use current frame feature map, $f8_t$, as current information, and 
% apply 
\sh{the} previous score map, $S_{t-n}$\sh{,} which is produced from score generator for the previous information.
% After then 
\sh{Then}, both feature maps are concatenated together and provided as an input to the template matching module to be compared with the \textit{template}.
The template matching module produces dissimilarity feature map, $D_t \in \mathbb{R}^{c_f, H, W}$\sh{,} which has the same resolution as $f8_t$ with a channel size of $c_{f}$.
The $D_t$ is forwarded to the reuse gate \sh{function} for deciding whether to skip \sh{the computation} or not.
If the reuse gate is on, the delta-generator makes a delta map $\Delta_t$ from $D_t$, which represents which pixels are changed from background to foreground and vice versa.
The delta-generator consists of 
% only 
a single convolution layer, so the computation is not heavy.
The loss, $Loss_{\Delta}$, accomplishes the above mentioned process by reducing the gap between $y'_t -S_{t-n}$ and $\Delta_t$ as follows:
\begin{equation}
\label{eq:Loss_D}
Loss_{\Delta} = L2(\Delta_t, y_t' - S_{t-n}).
\end{equation}
% Here, 
where $y_t'$ denotes the reduced sized ground truth mask at frame $t$ and $S_{t-n}$ is the previous score map. 
L2 loss minimizes the pixel-wise difference between $\Delta_t$ and $y_t'-S_{t-n}$.

In the initialization stage, the \textit{template} is generated with the given initial image $I_0$ and the corresponding mask $y_0$.
We reduce the resolution of the $y_0$ and consider the down-sampled mask as an initial score map $S_0$, i.e. $S_0 = y_0'$.
$f8_0$ is produced from $I_0$ and concatenated with $S_0$ to construct the \textit{template} as in TTVOS \cite{park2020ttvos}. 
Unlike TTVOS where the \textit{template} is updated every frame, our model skips the updating process to increase the speed of inference.

\noindent
\textbf{Applying to general VOS frameworks: }
In general VOS frameworks, the proposed method of template matching 
% for estimating object's movement 
can be applied by concatenating the current feature map and the estimated previous mask.

%% file: Texs/Method3.tex
\subsection{Estimation for Refined Feature Map}
\label{sec:refine}
When the reuse gate is on, our model skips layers of a segmentation network.
The segmentation network generates
% of base model generates 
the final accurate mask with the coarse score map $S$ along with features from the feature extractor.
% Owing to 
Using the delta-generator, our model can skip 
% partial 
layers of feature extraction step and generate $\hat{S}_t$ by adding $\Delta_t$ to the previous score map $S_{t-n}$.
Therefore, our model is unable to use the same segmentation network as the original FRTM.
Here, we explain how to estimate $\hat{R}_t$ using $R_{t-n}$ and $\Delta_t$ (See Fig. \ref{fig:arch}).

When the reuse gate is on, our model starts the segmentation network from $f8_t$.
However, the original network increases the resolution starting from $f32_t$, which is a feature map $32$ times smaller than the original input image size. 
% \hj{Due to different starting of the resolution of feature map,} 
\sh{Since we input different sized feature map to the network,} we design a refine-translator to estimate $\hat{R8}_t$ using $R8_{t-n}$ and $\Delta_t$.
Many segmentation networks used multi-sized receptive fields to improve the accuracy \cite{mehta2018espnet, park2018c3, park2020sinet}.
We 
% also 
adopt this method with different dilated ratios.
As shown in Fig \ref{fig:RefineT}(a), $R8_{t-n}$ and $\Delta_t$ are concatenated and passed on to different dilated convolutions.
Each output feature map has to embed features with different receptive fields to cope with various object sizes.
Finally, the entire information is merged using a 
% single 
convolution layer and a 
% single 
ResBlock.
After the refine-translator, the remaining process is the same as that of the original FRTM.
Fig. \ref{fig:RefineT}(b)-(e) explains the overall process with two consecutive frames, 24 and 25, in a \textit{goat} video.
Fig. \ref{fig:RefineT}(d) shows $\hat{R8}_{25}$ in the first row and $R8_{24}$ in the second row.
$\Delta_{24}$ is depicted in the first row of (f). The second row of (e) shows the generated mask 
% which is 
produced using $\hat{S}_{25}$, $f8_{25}$, $f4_{25}$ and $\hat{R8}_{25}$.

% Bottom of Fig. \ref{fig:RefineT} (d) is $\hat{R8}_{28}$ from $R_{27}$ which is top of Fig. \ref{fig:RefineT} (d) and $\Delta_{27}$ which is top of Fig. \ref{fig:RefineT} (f).
% Finally, the model produces accurate target mask with $\hat{S}_{28}$, $f8_{28}$, \sh{and} $f4_{28}$ as shown \sh{at the} bottom of Fig. \ref{fig:RefineT} (e).

%% file: Texs/Method4.tex
\subsection{Gate Probability Loss}
\label{sec:gating}
% The fixed constraint on the number of gates that can be used in the computation is inappropriate in our setting, \jy{our goal is not to skip the layers of the network in fixed ratio, but to skip the layers only when the consecutive frames are similar}. % wish->intend
% We \hj{intend} 
% % to learn 
% to set the reuse gate on (reuse the previous frame's feature) when the previous frame $t-n$ and the current frame $t$ are substantially similar.
% For this purpose, we propose a gate probability loss that is based on IoU as follows:
% \begin{equation}
% \label{eq:iou}
% \begin{split}
% P_{target} = Max(m_1,\; IoU^{t-n}_t) , \\
% % P_{target}=[p,\; 1-p] 
% \end{split}
% \end{equation}
% \begin{equation}
% \label{eq:loss_GP}
% Loss_{gp} = Max(m_2,\;|P_{gate}- P_{target}|)^2 
% \end{equation}

% As mentioned in Sec \ref{sec:arch}, output of the gate function is the probability of the gate being on and off.

% This section explains how we train our dynamic network. The challenge is twofold: Firstly, we need to train multiple network paths as well as the gating logic (reuse gate). The second challenge is to encourage the network to achieve high segmentation accuracy while choosing the cheaper computation path as often as possible.
This section explains how we train our dynamic network. 
% The challenge is twofold: Firstly, we need to train not fixed number of selection but multiple number of selection for the gating logic (reuse gate) depending on input frames adaptively.
The challenge is twofold: Firstly, we need to train multiple network paths as well as the gating logic (reuse gate). 
In the detail, the quantity of target’s movement varies depending on consecutive input frames in the training stage.
Therefore, the optimal selection of reuse gate is diverse for each iteration.
We desire that if the quantity of movement is a lot, model learns not to select the reuse gate, while if the quantity is little, model learns to select the reuse gate.
The second challenge is to encourage the network to achieve high segmentation accuracy while choosing the cheaper computation path as often as possible.

To accomplish this, we use the following training recipe -- when the model training begins, we train the reuse gate function to predict the IoU of the current frame's ground truth mask with respect to that of the previous frame.
Based on $P_{gate}$, as mentioned in Eq. (\ref{eq:gate}) of Sec \ref{sec:arch}, we train the different paths for mask generation.
As the training progresses, we want to avoid the situation where the network predicts a low value for $P_{gate}$ to select the more expensive full mask generation path for achieving a higher IoU score.
We accomplish this by artificially boosting the IoU between the current and the previous frame's mask introducing a margin which is gradually increased until it reaches $m_1$ from $0$.
By doing so, we direct the network to select the skip path more often and achieve higher accuracy even with the reduced computation.

To realize the above training schedule, we propose a novel gate probability loss that is based on IoU as follows:
\begin{equation}
\label{eq:iou}
\begin{split}
m=m_1*(ep_c/ep_T), \\
P_{target} = Max(m,\; IoU^{t-n}_t), \\
% P_{target}=[p,\; 1-p] 
\end{split}
\end{equation}
\begin{equation}
\label{eq:loss_GP}
Loss_{gp} = Max(m_2,\;|P_{gate}- P_{target}|)^2 
\end{equation}
% The reuse gate function outputs the probability of the gate being on (reuse), $P_{gate}$.
% This probability is proportional to the overlapped region between ground truth mask frames at time $t-n$ and $t$.
where $IoU^{t-n}_t$ is IoU between the ground truth mask frames at time $t-n$ and $t$, and $ep_c$ is the current epoch and $ep_T$ is the target epoch at which time $m$ reaches to $m_1$. 
We set 120 for $ep_T$.
% To induce the stronger constraint of reusing previous frame's information to the model, we applied a margin $m_1$, which is gradually increased until it reaches 1.
% Therefore, the target probability of the reuse gate being on is increased and the probability of the reuse gate being off is decreased in $P_{target}$ gradually.
A gate probability loss, denoted by $Loss_{gp}$, penalizes the model proportionally to the difference between $P_{gate}$ and $P_{target}$, when the difference is larger than the margin $m_2$.
The final loss becomes:
\begin{equation}
\label{eq:loss}
Loss = Loss_{gp} + Loss_{\Delta} + BCE(y_t, \hat{y_t}), 
\end{equation} 
where $BCE$ is \nj{the} binary cross entropy loss between the pixel-wise ground truth $y_t$ at frame $t$ and its estimation $\hat{y}_t$.

% This section explains how we train our dynamic network. The challenge in training this dynamic architecture is twofold: Firstly, we need to train multiple network paths as well as the gating logic (reuse gate). The second challenge is to encourage the network to achieve high segmentation accuracy while choosing the cheaper computation path as often as possible.

% To accomplish this, we use the following training recipe -- when the model training begins, we train the reuse gate function to predict the IoU of the current frame's mask with respect to the previous frame's ($P_{gate}$). Based on $P_{gate}$, we train the different paths for mask generation. As the training progresses, we want to avoid the situation where the network predicts a low value for $P_{gate}$ to select the more expensive full mask generation path for achieving a higher IoU score. We accomplish this by artificially boosting the IoU between the current and previous frame's mask. By doing so, we direct the network to select the compute skip path more often and achieve higher accuracy even with this reduced computation.

% To realize the above training schedule, we propose a novel gate probability loss that is based on IoU as follows:
% \begin{equation}
% \label{eq:iou}
% \begin{split}
% P_{target} = Max(m_1,\; IoU^{t-n}_t) , \\
% % P_{target}=[p,\; 1-p] 
% \end{split}
% \end{equation}
% \begin{equation}
% \label{eq:loss_GP}
% Loss_{gp} = Max(m_2,\;|P_{gate}- P_{target}|)^2 
% \end{equation}

%% file: Texs/Exp1.tex
\section{Experiment}
\label{sec:exp}

In this section, we prove the efficacy of the proposed method using the official benchmark code of DAVIS~\cite{perazzi2016benchmark, Pont-Tuset_arXiv_2017}\footnote{{https://github.com/davisvideochallenge/davis2017-evaluation}} and the official evaluation server of YouTube-VOS 2018~\cite{xu2018youtube}\footnote{{https://competitions.codalab.org/competitions/19544}}.
DAVIS 16 is a single object task that consists of $30$ training videos and $20$ validation videos, while DAVIS 17 is a multiple object task with $60$ training videos and $30$ validation videos.
The average of mean Jaccard index $J$ and F-measure $F$ are used to report the model accuracy in the DAVIS benchmark.
$J$ index measures the overall similarity by comparing estimated masks with ground truth masks and $F$ score focuses on the boundary by delimiting the spatial extent of the mask.
YouTube-VOS is the largest VOS dataset that consists of 3,471 training videos and 474 validation videos (in total 4,453 videos). 
The validation set is split into seen (65 categories) and unseen (26 categories) to evaluate the generalization ability.

We compare our model with other state-of-the-art models by extending it to two VOS models: FRTM and TTVOS. 
% and show further application of the proposed method on a different architecture, TTVOS.
Our ablation study shows that the refine-translator and the gate probability loss are important factors in preserving the original accuracy and in activating gate properly.
We measure detailed performance degradation with and without the refine-translator using different values of $\tau$ in Eq. (\ref{eq:switch}).
Furthermore, we report performance changes depending on the different setting of margins, $m_1$ and $m_2$, in Eq. (\ref{eq:iou}).

\noindent
\textbf{Implementation Details: }
We implement our method with the FRTM official code~\footnote{{https://github.com/andr345/frtm-vos}}.
FRTM consists of two versions: FRTM and FRTM-fast.
FRTM uses ResNet101 and FRTM-fast uses ResNet18 for feature extraction, and different numbers of iterations are used for fine-tuning the score generator.
We follow their training scheme with the following modifications to better fit out dynamic architecture training.
We change batch size from 16 to 8 and increase the number of sequences from 3 to 6 to train with greater temporal history within the same memory budget.
% When the number of sequences is 3, because the model uses the first frame for the initialization step and the rest for the training of the model, our model cannot learn from a long range of similarity.
The learning rate is decreased from $1e^{-3}$ to $5e^{-4}$ and we use the total training epoch of 260.
Following the setting of AIG~\cite{veit2018convolutional}, we initialize the reuse gates to be on with the probability of 15\%.

\subsection{DAVIS Benchmark Result}
\label{sec:davis}

%%%%%%%%%%%%%%%%%%%%%%%%%%%% Main figure of experiment %%%%%%%%%%%%%%%%%%%%%%%%%%%%%%%%%%%%%%
\begin{figure}[t]
\begin{center}
  \includegraphics[width=0.9\linewidth]{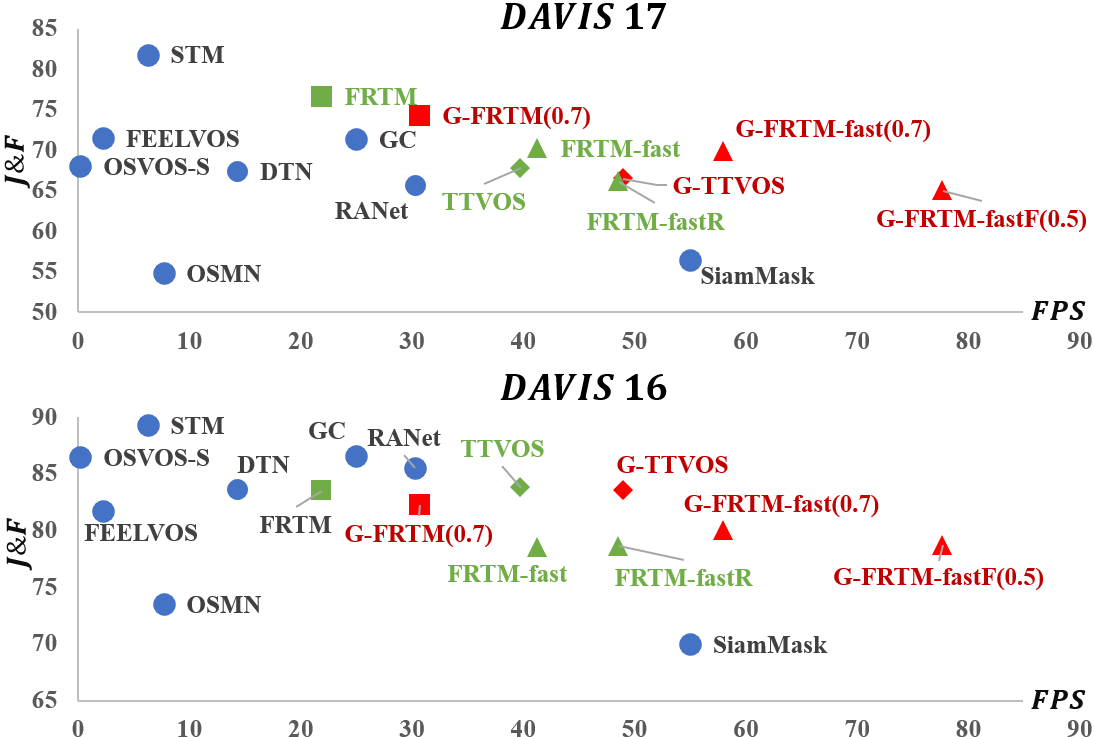}
\end{center}
  \caption{FPS vs $J\&F$ score on the DAVIS validation sets.  
  $\triangle$, $\square$, and $\diamond$ denotes experiments based on FRTM-fast, FRTM, and TTVOS, respectively.
   \textbf{G-} indicates using the proposed method and fastF denotes experiments based on FRTM-fast with the fusion method and FRTM-fastR is result of reducing a channel size from multiple layers in original FRTM-fast as shown on Tab.~\ref{tab:abl}.
  }
\label{fig:performance}
\end{figure}
%%%%%%%%%%%%%%%%%%%%%%%%%%%% Main figure of experiment %%%%%%%%%%%%%%%%%%%%%%%%%%%%%%%%%%%%%%

\begin{table}[t]
\scriptsize
  \centering
 \setlength\tabcolsep{3.5pt} 
    \begin{tabular}{l l  |ccc|cc|c}
   \Xhline{3\arrayrulewidth}
                                   &   &\multicolumn{3}{c|}{Train Dataset}  &  \multicolumn{2}{c|}{DAVIS}  &  \\
    Method                                  & Feature & Ytb    & Seg   & Syn&  17  &   16   &  FPS\\
    \hline
    OnAVOS \cite{DBLP:conf/bmvc/VoigtlaenderL17}  & VGG16        & - & o & - & 67.9    & 85.5  & 0.08 \\
    OSVOS-S \cite{maninis2018video}               & VGG16        & - & o & - & 68.0    & 86.5  & 0.22 \\

    STM \cite{oh2019video}                        & RN50     & o & - & o & 81.8  & 89.3  & 6.25 \\
    GC \cite{li2020fast}                          & RN50     & o & - & o & 71.4  & 86.6  & 25.0 \\
   
    OSMN \cite{yang2018efficient}                 & VGG16        & - & o & - & 54.8  & 73.5  & 7.69\\
    RANet \cite{wang2019ranet}                    & RN101    & -  & - & o & 65.7  & 85.5  & 30.3 \\
    A-GAME \cite{johnander2019generative}         & RN101    & o & - & o & 70    &   82.1    & 14.3 \\
    FEELVOS \cite{voigtlaender2019feelvos}       & XC65   & o & o & - & 71.5  & 81.7  & 2.22 \\
    SiamMask \cite{wang2019fast}                 & RN50      & o & o & - & 56.4  & 69.8    & 55.0 \\ 
    DTN \cite{Zhang_2019_ICCV}                  & RN50      & - & o & - & 67.4  & 83.6    & 14.3 \\
    FRTM \cite{robinson2020learning}             & RN101    & o & - & - & 76.7  & 83.5  & 21.9 \\
    
    FRTM-fast \cite{robinson2020learning}        & RN18      & o & - & - & 70.2  & 78.5  & 41.3 \\
    % TTVOS \cite{park2020ttvos}                                  & HRN         &  o &  - & o & 58.7& 79.5 & 73.8 \\
    TTVOS \cite{park2020ttvos}                               & RN50         &  o &  - & o & 67.8& 83.8 & 39.6 \\
    \hline
    \hline
    G-FRTM              ($\tau=1$)                 & RN101       & o & - & - & 76.4  & 84.3  & 18.2 \\
    G-FRTM              ($\tau=0.7$)              & RN101       & o & - & - & 74.3  & 82.3  & 28.1 \\
    G-FRTM-fast         ($\tau=1$)                 & RN18     & o & - & - & 71.7  & 80.9  & 37.6 \\
    G-FRTM-fast         ($\tau=0.7$)               & RN18      & o & - & - & 69.9  & 80.5  & 58.0 \\
    % G-TTVOS [cite]      ( TH=0 )                   & RN50     &  o &  - & o & 65.5& 82.9 & 39.0 \\
    G-TTVOS       ($\tau=0.7$)                & RN50      &  o &  - & o & 66.5& 83.5 & 49.1 \\
    % G-TTVOS (TH=0.45) [cite]                    & RN50      &  o &  - & o & 63.3& 78.1 & 52.1 \\
    \Xhline{3\arrayrulewidth}
    \end{tabular}%
    
     \caption{Quantitative comparison on the DAVIS benchmark validation set. 
     Ytb represents using Youtube-VOS for training.
     Seg is a segmentation dataset for pre-training by Pascal \cite{everingham2015pascal} or COCO \cite{lin2014microsoft}.
     Syn is using a saliency dataset for making synthetic video clip by affine transformation. 
     RN, and XC denotes ResNet and Xception for feature extraction, respectively.
     \textbf{G-} indicates using the proposed method based on FRTM and TTVOS.
     %Similar to other works, 
    %  FPS is measured on DAVIS16. hj: Last time one reviewer tackle that "do not compare the accuracy in DV17"
    \hj{Similar to other works, we measure FPS on DAVIS 16.}
    Note that performances and speed on baseline models are taken from original papers.
     } 
  \label{tab:davis}%
\end{table}%

We compare our method with other state-of-the-art models, as shown in Tab. \ref{tab:davis} and Fig.~\ref{fig:performance}.
We report a model used for feature extraction and training datasets for clarification, since each model has a different setting.
Furthermore, we also show additional results on TTVOS to claim the generality of our method.
Our method improves the inference speed without any significant accuracy degradation.
In Tab.~\ref{tab:davis}, a prefix of \textbf{G-} denotes the implementation of the proposed method on the baseline models, FRTM and TTVOS.
In the slowest case of $\tau=1$, the model uses every layer of the network, which is equivalent to using the original baseline model.
Our model reports slightly higher performance when applied to FRTM with $\tau=1$ than the original FRTM.
We assume that the difference comes from the different number of training sequences used, as in our implementation 6 is used instead of 3.
In case of $\tau=0.7$, the average reuse rate in the model is $0.402$ for DAVIS 17 and $0.475$ for DAVIS 16 with marginal performance degradation of $1.8$ and $0.4$. % , respectively.
FPS improves from $24.6$ to $37.8$ and $37.6$ to $58.0$ on each dataset, which are $1.5$ times faster than the slowest case of $\tau=1$.
% Some VOS models does not generate coarse score map for the template matching. 
% We also apply our method to TTVOS, one of architectures that does not produce the score map, and show that our method can bring improvement of speed over the baseline model.
% Details of the architecture are explained in the supplementary material.
We also apply our method into TTVOS to prove that our method can be used for other VOS models regardless of whether a model produces the score map or not.
Successfully, our method can bring improvement of speed over the baseline model without performance degradation, and details of the architecture are explained in the supplementary material.
Moreover, our dynamic method is much faster than DTN~\cite{Zhang_2019_ICCV}, a dynamic network with switching modules, with better accuracy on DAVIS 17 and comparable accuracy on DAVIS 16.

%% file: Texs/Exp2.tex
\subsection{Ablation Study}
\label{sec:ablation}

\begin{figure}[t]
\begin{center}
%   \fbox{\rule{0pt}{1.5in} \rule{0.9\linewidth}{0pt}}
 \includegraphics[width=0.9\linewidth]{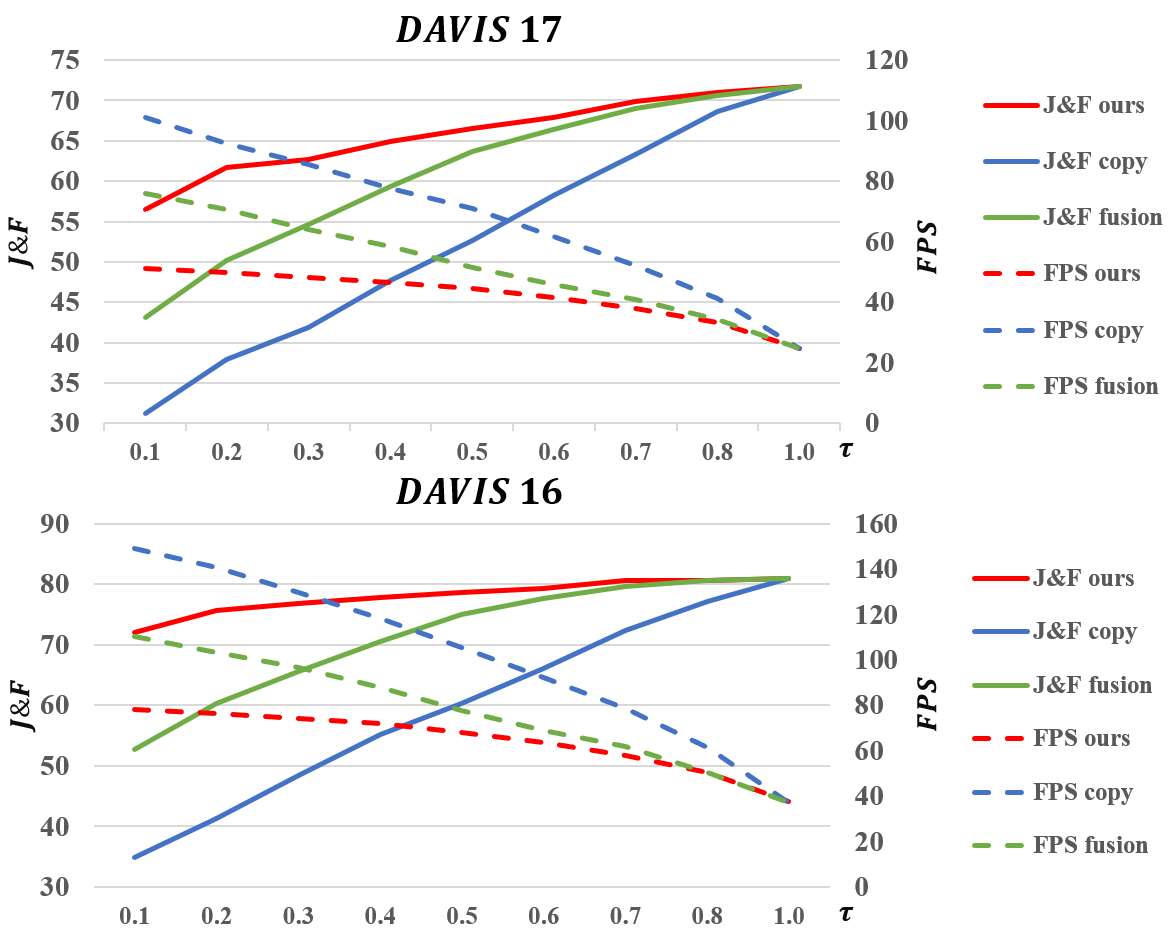} 
\end{center}
   \caption{Ablation study about different method for reusing previous information, when the reuse gate is on by comparison of accuracy and FPS on DAVIS with different $\tau$.
    Ours is our method based on FRTM-fast.
    \textbf{Copy} simply copies the previous mask for the current frame, without using the refine-translator.
    \textbf{Fusion} copies the previous mask if the similarity of consecutive frames is extremely large. Otherwise, original method of the refine-translator is used. } % comment_sh: what about 'similarity is high' instead of 'dissimilarity is small'? 
\label{fig:abl_refine}
\end{figure}

\begin{table}[t]
  \centering
  \footnotesize
   \begin{tabular}{l | cc | ccc}
   \Xhline{3\arrayrulewidth}
                & $Loss_{gp}$ & D/R       & dv17      & dv16     & FPS \\
    \hline
    
    FRTM-fast                & x     & x     & 70.2  & 78.5  & 41.3 \\
    FRTM-fast*                &  x    & x     & 71.7  & 81.3  & 40.1 \\
    FRTM-fastR                &  x    & x     & 66.1 & 78.6  & 48.6 \\
    $Loss_{Ngate}$   & x     & o     & 61.3  & 76.5 & 37.8 \\
    \hline
    Ours-copy ($\tau=0.7$)           & o     & x     & 63.3  & 72.4 & 75.6 \\
    Ours-copy ($\tau=0.5$)           & o     & x     & 52.6 & 60.3 & 100.8 \\
    Ours-copy ($\tau=0.1$)           & o     & x     & 31.2 & 34.8 & 150.2 \\
    \hline
    Ours-fusion ($\tau=0.7$)       & o   & $\triangle$  & 69.0 & 79.6 & 61.8 \\
    Ours-fusion ($\tau=0.5$)       & o   & $\triangle$  & 63.7 & 75.0 & 77.7 \\
    \hline
    Ours ($\tau=1$)                & o     & o     & 71.7  & 80.9  & 37.8 \\
    Ours ($\tau=0.7$)                & o     & o     & 69.6  & 80.5  & 58.0 \\
    Ours ($\tau=0.5$)                & o     & o     & 66.5  & 78.7  & 68.2 \\
    Ours ($\tau=0.1$)                & o     & o     & 56.5  & 72.0 & 78.1 \\

    \Xhline{3\arrayrulewidth}
    \end{tabular}%

    \caption{Ablation study on the proposed modules and the loss function.
     $Loss_{Ngate}$ means training the gate function using the constraint of the number of gates being on.
     $D/R$ means using the delta-generator and the refine-translator, in the case of reuse gate being on.
     % copy method -> original -> full path
    %  $\triangle$ indicates using the copy method when the estimated similarity has a high value.
      $\triangle$ indicates using one of \textbf{copy}, ours, and the full path calculation based on the value of the estimated similarity.
     FRTM-fast* is our implementation with the same training schemes as our gate function.
    %  Note that compared to the original implementation of FRTM-fast, our implementation shows slightly higher performance with slightly slower inference speed due to different experiment environment.
     FRTM-fastR is our implementation of reducing the number of channels in multiple layers to make its inference speed similar to ours.
     }
  \label{tab:abl}%
\end{table}%

\begin{figure}[t]
\begin{center}
%   \fbox{\rule{0pt}{1.5in} \rule{0.9\linewidth}{0pt}}
 \includegraphics[width=0.87\linewidth]{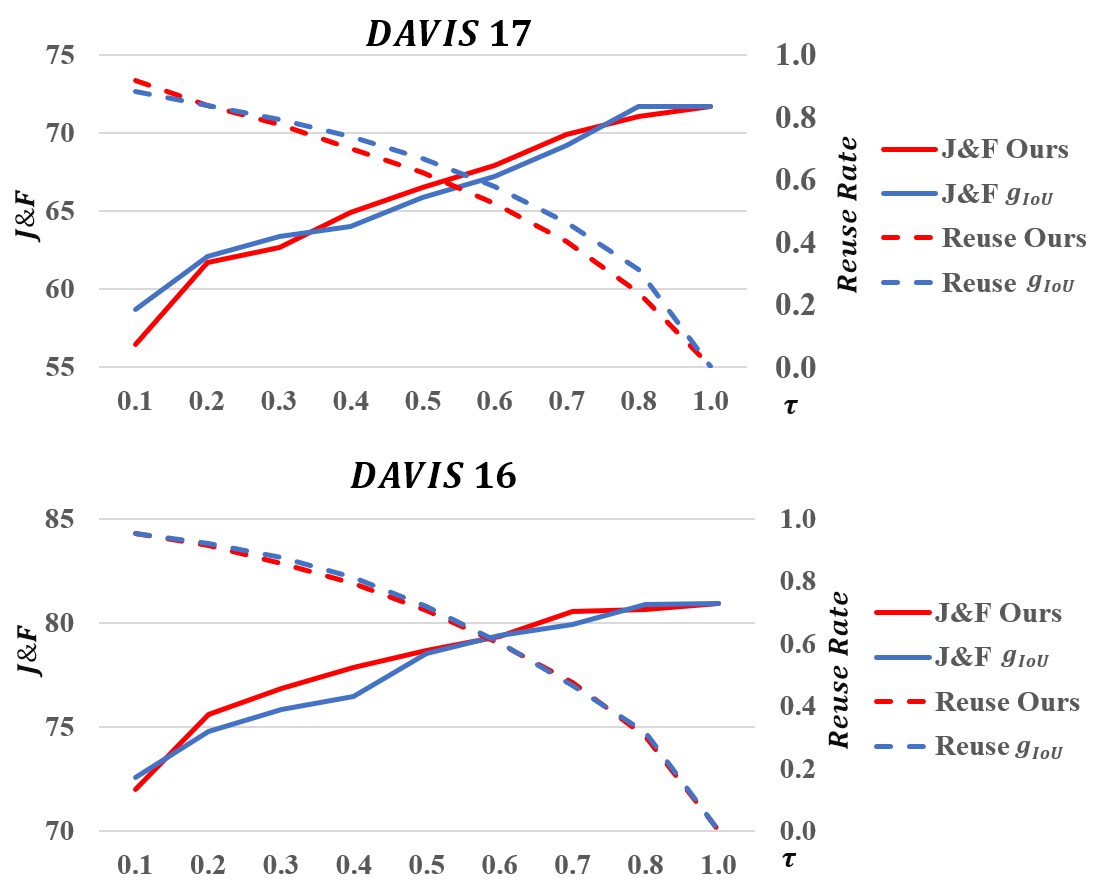}
\end{center}
   \caption{Ablation study about $P_{gate}$%methods for measuring similarity
   by comparison of accuracy and reuse rate on DAVIS with different setting of $\tau$ .
   Ours estimates the similarity by gate function for deciding gate being on or off.
   $g_{IoU}$ is using ground truth IoU between adjacent frames as a similarity for deciding gate.}
\label{fig:abl_g}
\end{figure}

% comparison of the output of the gate function and the IoU between the previous and the current ground truth masks depending on different $\tau$ values. The gate function outputs the probability of similarity.
% gating function(D_t) -> prob of dissimilarity. dis prob의 gt는 previous랑 current의 1-IoU. 

In this section, we analyze our modules to show the importance of using 1) \sh{the} refine-translator, 2) \sh{the} gate probability loss and 3) \sh{the} margin in gate probability loss for preserving original accuracy.

Tab.~\ref{tab:abl} and Fig.~\ref{fig:abl_refine} demonstrate \sh{the} effect of \sh{the} refine-translator.
Originally, the refine-translator takes the previous refined feature map and estimates a feature map corresponding to the current stage.
To validate the importance of the refine-translator, we 
% conduct
\sh{implement}
% another 
\sh{other} method\sh{s} to replace original method with \textbf{copy} and \textbf{fusion}\sh{.}
\textbf{copy} indicates that the model simply copies the previous mask as a result of the current frame when the reuse gate is set on. Therefore, these models do not use the refine-translator.
\textbf{fusion} is a mixed method between the \textbf{copy} and the original method with additional threshold of $\tau_2$ which is greater than $\tau$.
When the reuse gate is on and the probability value is greater than $\tau_2$, the model copies the previous mask for the current frame due to extreme similarity.
As shown in Fig. \ref{fig:abl_refine}, models that use \textbf{copy} method experience significant performance degradation, while models with our method manage to preserve the original accuracy.
In our method, when $\tau=0.1$, the reuse rate becomes 95.4\% on DAVIS 16, with minimal $9\%$ of performance degradation and the inference speed becomes twice faster than when $\tau=1$.
However, in case of the \textbf{copy} method, the accuracy decreases by $46\%$.
The \textbf{fusion} method takes both advantages from the original and \textbf{copy} methods.
The performance and speed of \textbf{fusion} are drawn in the middle of the our method and the \textbf{copy} method.
It suggests an adequate alternative to the proposed method when greater increase in inference speed is needed.
% For example, when $\tau=0.5$ in the \textbf{fusion} method, the speed and the accuracy are better than TTVOS on DAVIS17, as shown in Fig.~\ref{fig:performance}. \sh{\textbf{Q:} what's the purpose of this last sentence?}

\begin{figure}[t]
\begin{center}
%   \fbox{\rule{0pt}{1.5in} \rule{0.9\linewidth}{0pt}}
 \includegraphics[width=0.9\linewidth]{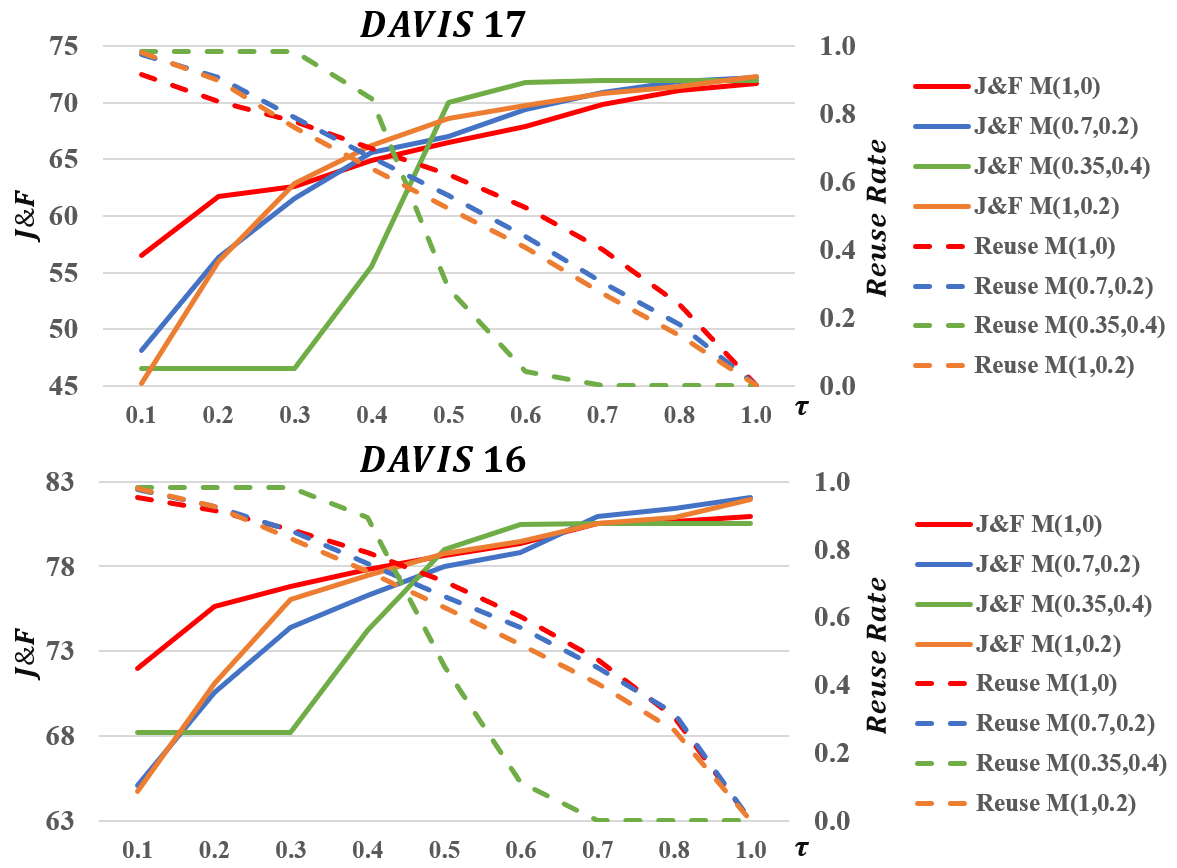} 
\end{center}
   \caption{Comparison of accuracy and reuse rate on DAVIS with different settings of margin in the gate probability loss.
   We used $M(1,0)$ for training. }
\label{fig:abl_M}
\end{figure}

Tab.~\ref{tab:abl} 
% also 
shows \sh{the} importance of using the gate probability loss function for 
% maintaining 
\sh{preserving} original accuracy.
The experiment of $Loss_{Ngate}$ is the result of using a fixed constraint on the number of gates that can be used in the computation as 
% like 
\sh{in} \cite{lee2020urnet,veit2018convolutional}.
% There is huge performance degradation, 
\sh{Using $Loss_{Ngate}$ shows huge degradation,} even \sh{when} the module does not reuse any \sh{of the} previous features. Furthermore, our method with $\tau=0.7, \,0.5$ shows better performance and FPS compared to the case of FRTM-fastR, where channels of multiple layers in FRTM-fast, except a feature extractor, are reduced to meet the similar inference speed as ours.
Fig. \ref{fig:abl_g} demonstrates that our gate function is working properly following the ground truth similarity (IoU) between the current and the previous masks.
\textbf{$g_{iou}$} uses ground truth IoU instead of the estimated similarity probability from the gate function to decide whether to turn the reuse gate on or not.  
Our results on reuse rate and accuracy coincide with when \textbf{$g_{iou}$} is used. % instead.

Finally, Fig.~\ref{fig:abl_M} describes \sh{the} effect of different settings of margin $M(m_1,m_2)$ in the gate probability loss as mentioned in Sec.~\ref{sec:gating}.
We conduct experiments on various settings, $m_1=0.35, 0.7, 1.0$, and $m_2=0.0, 0.2, 0.4$. M(1,0) is \sh{used in} our setting.
When the value of $\tau$ is large, the performance of each experiment shows similar trend. However, when the value approaches to 0, which forces the model to reuse the previous information more, the accuracy of $M(1,0)$ is much higher than others.
$M(1,0.2)$ and $M(0.7,0.2)$ have similar accuracy and reuse rate.
$M(0.35,0.4)$ works improperly.
When $\tau$ changes from $0.6$ to $0.5$, the model suddenly decides to reuse $80\%$ of the frames. % with $M(0.35,0.4)$.
Therefore, we think $m_2$, which is a margin for \sh{the} gap between $P_{gate}$ and $P_{target}$, is a more important factor for preserving the accuracy.

\subsection{YouTube-VOS Result }
\label{sec:ytb}

\begin{table}[t]
\footnotesize
  \centering
   \setlength\tabcolsep{2.5pt} 
    \begin{tabular}{ll |cc|ccccc}
      \Xhline{3\arrayrulewidth}
      
              &       &\multicolumn{2}{c|}{Dataset} & $G$ & \multicolumn{2}{c}{$J$} & \multicolumn{2}{c}{$F$} \\
    Method                           & Ft             & seg  & syn        & All & S  & Us & S  & Us \\
    \hline
    onAVOS\cite{DBLP:conf/bmvc/VoigtlaenderL17}      & VG & o  & -  & 55.2  & 60.1  & 46.1  & 62.7  & 51.4 \\
    OSVOS\cite{robinson2020learning}                 & VG & o  & -  & 58.8  & 59.8  & 54.2  & 60.5  & 60.7\\
    S2S \cite{xu2018youtube}                        & VG & -  & -  & 64.4  & 71.0  & 55.5  & 70.0  & 61.2 \\
    PreMVOS \cite{Luiten18ECCVW}                  & RN*  & o  & o  & 66.9  & 71.4  & 56.5  & -     & - \\
    STM \cite{oh2019video}                          & R50 & -  & o & 79.4  & 79.7  & 72.8  & 84.2  & 80.9 \\
    GC \cite{li2020fast}                            & R50 & -  & o & 73.2  & 72.6  & 68.9  & 75.6  & 75.7 \\
    A-GAME \cite{johnander2019generative}           & R101 & -  & o & 66.1  & 67.8  & 60.8  & 69.5  & 66.2 \\
    RVOS \cite{Ventura_2019_CVPR}                   & R101 & -  & - & 56.8  & 63.6  & 45.5  & 67.2  & 51.0 \\
    FRTM-fast \cite{robinson2020learning}           & R18  & -  & -  & 65.7  & 68.6  & 58.4  & 71.3  & 64.5 \\
    \hline
    FRTM-fast*                      & R18 & -  & o & 61.9  & 67.0  & 52.6  & 69.5 & 58.6 \\
    G-FRTM-fast ($\tau=1$)              & R18 & -  & o & 60.9  & 65.1  & 53.0  & 66.7 & 58.8 \\
    G-FRTM-fast ($\tau=0.6$)          & R18 & -  & o & 60.3  & 64.3  & 53.1  & 65.2  & 58.6 \\
    % G-FRTM-fast+ ($\tau=1$)              & R18 & -  & o & 60.9  & 65.1  & 53.0  & 66.7 & 58.8 \\
    % G-FRTM-fast+ ($\tau=0.6$)          & R18 & -  & o & 60.3  & 64.3  & 53.1  & 65.2  & 58.6 \\
    % G-FRTM ($\tau=1$)                   & R101 & -  & - & 56.8  & 63.6  & 45.5  & 67.2  & 51.0 \\
    
  \Xhline{3\arrayrulewidth}
    \end{tabular}%
    \caption{Quantitative comparison on YouTube-VOS benchmark validation set. 
     Seg is a segmentation dataset for pre-training by Pascal \cite{everingham2015pascal} or COCO \cite{lin2014microsoft}.
     Syn is saliency datasets for making synthetic video clips by affine transformation.
     S and Us are seen and unseen categories. 
     VG is VGG16 and R is ResNet.
     RN* is a variation of ResNet proposed in \cite{ovsep2017large}.
     FRTM-fast* is our implementation with \nj{the} same training schemes as our gate function.
      \textbf{G-} indicates using proposed method based on FRTM-fast. 
      Note that performances on baseline models are taken from original papers.
     }
  \label{tab:ytb}%
\end{table}%

\begin{table}[t]
  \centering
\footnotesize
\setlength\tabcolsep{4.5pt} 
    \begin{tabular}{lc|ccccc}
    \Xhline{3\arrayrulewidth}
          &       & G     & \multicolumn{2}{c}{J} & \multicolumn{2}{c}{F} \\
    \multicolumn{1}{c}{Method} & reuseR & All   & S     & Us    & S     & Us \\
    \hline
    G-FRTM-fast ($\tau=1$) & 0     & 63.8  & 68.3  & 55.2  & 70.6  & 61.0 \\
    G-FRTM-fast ($\tau=0.8$) & 25.5  & 63.4  & 67.6  & 55.8  & 69.3  & 60.9 \\
    G-FRTM-fast ($\tau=0.7$) & 40.0  & 62.7  & 67.1  & 55.2  & 68.2  & 60.1 \\
    G-FRTM-fast ($\tau=0.6$) & 50.9  & 62.3  & 66.7 & 55.3 & 67.2 & 60.0 \\
    \Xhline{3\arrayrulewidth}
    \end{tabular}%
  \caption{Quantitative comparison on YouTube-VOS benchmark validation set with different training scheme. 
    %  There are two option for validation. One is using all the frames in video and the other is using only sampled frames.
    %  Allf means using all the frames in the video.
    %  Ep is total number of epoch for training, and reU is reusing rate in whole sequences.
    \textbf{reuseR} denotes reusing rate.
     S and Us are seen and unseen categories. 
      \textbf{G-} indicates using proposed method based on FRTM-fast.      }
  \label{tab:ytb_supp}%
\end{table}%

%%% Qualitative example ###############################################

\begin{figure*}[t]
\begin{center}
\includegraphics[width=0.9\linewidth]{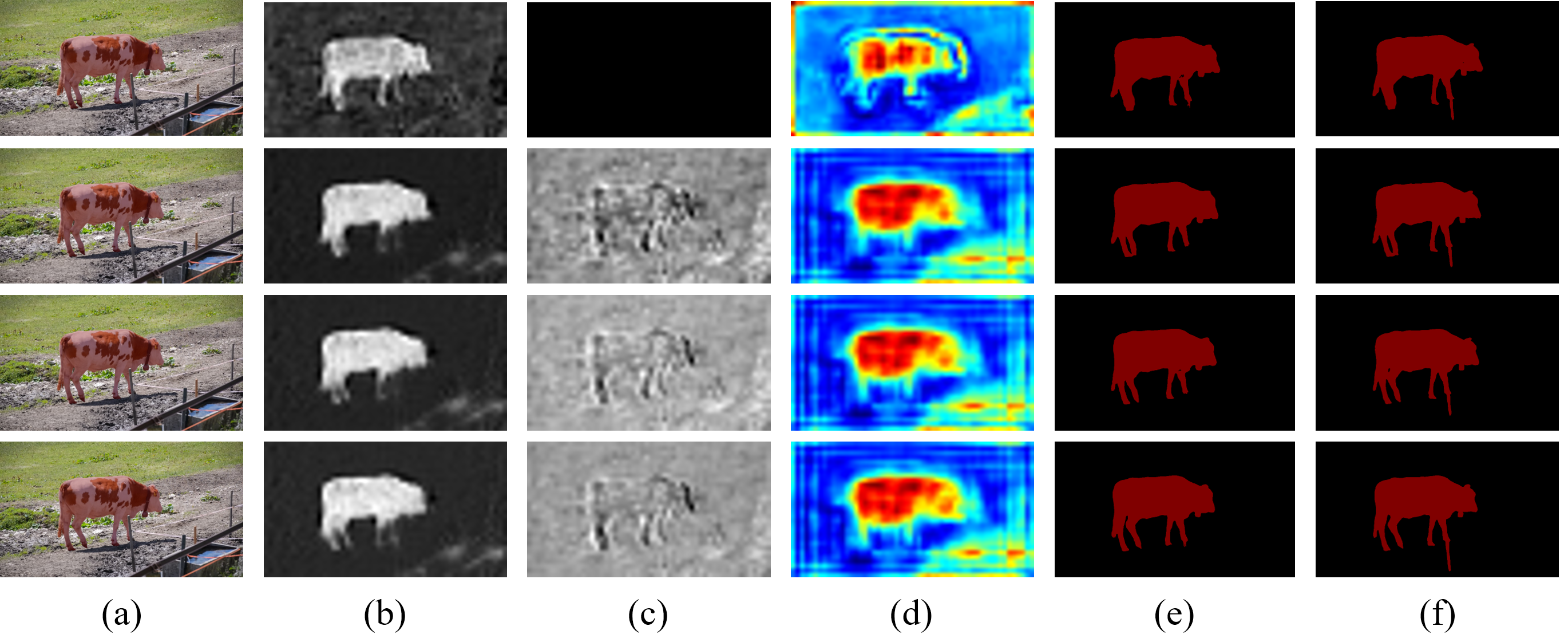}\\
\end{center}
   \caption{(a)-(f) Example of \textit{cows} frame $23-26$.
   (a) Input frames are overlapped with ground truth masks.
   (b) $S$ and $\hat{S}$. Top of row is $S$ and the others are $\hat{S}$.
   (c) $\Delta_t$. The black image of top of row means this frame is not reused. Therefore, the $\Delta_t$ is not generated
   (d) $R8_t$ and $\hat{R8}_t$. Top of row is $R8_t$ and the others are $\hat{R8}_t$.
   (e) Our results
   (f) FRTM-fast results
   }
\label{fig:Q_single}
\end{figure*}

\label{sec:Supple_fig}

\begin{figure*}[t]
\begin{center}
\includegraphics[width=0.9\linewidth]{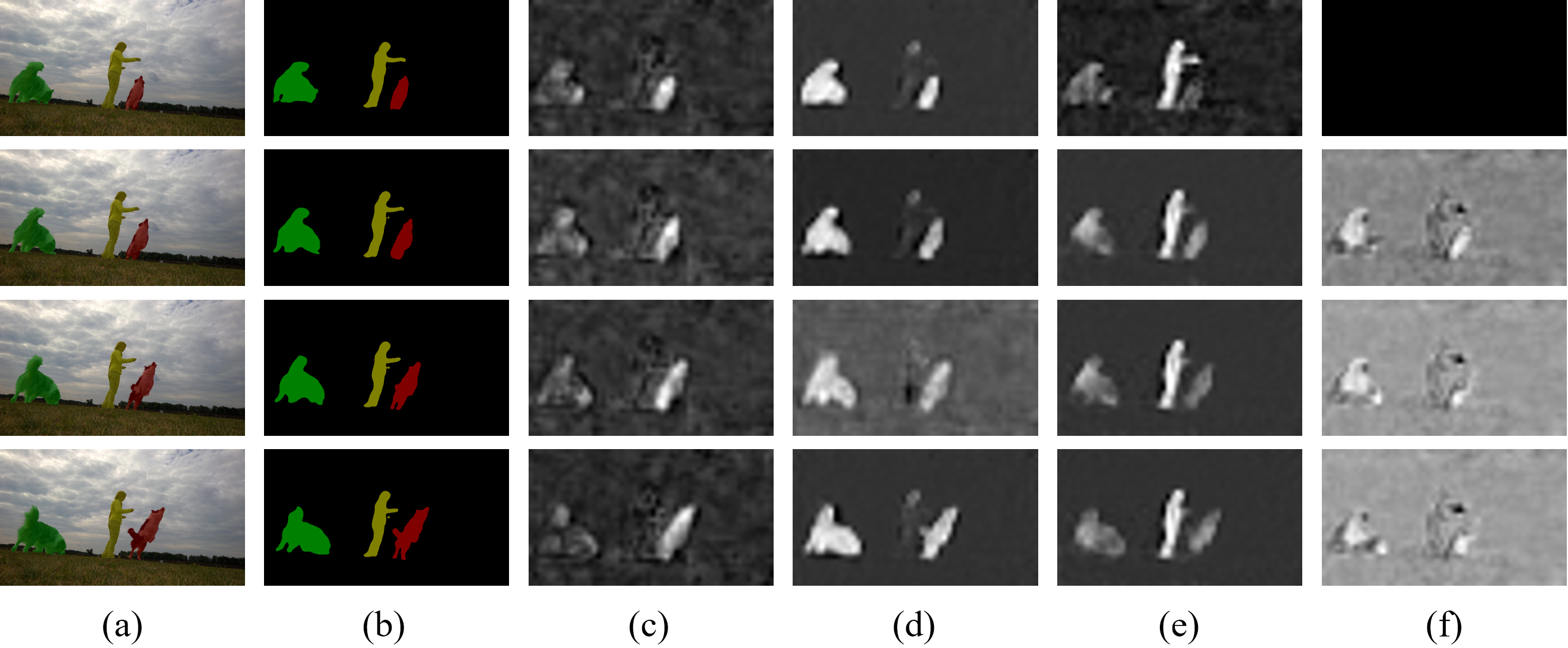}\\
\end{center}
   \caption{(a)-(f) Example of \textit{dogs-jump} frame $29-32$. The red dog is object1, the green dog is object2 and human is object3. The reuse rates are 0.152, 0.439 and 0.864, respectively.
   (a) Input frames are overlapped with ground truth masks.
   (b) Our results 
   (c) $S$ regarding to object1. All the frame does not reuse previous features
   (d) $S$ and $\hat{S}$ regarding to object2. Third of row is $S$ and the others are $\hat{S}$.
   (e) $S$ and $\hat{S}$ regarding to object3. Top of row is $S$ and the others are $\hat{S}$.
   (f) $\Delta_t$ regarding to object3. The black image of top of row means this frame is not reused.
   Therefore, the $\Delta_t$ is not generated
   }
\label{fig:Q_multi}
\end{figure*}

%%% Qualitative example ###############################################

Tab. \ref{tab:ytb} shows our result on YouTube-VOS dataset.
FRTM-fast* is a result of our implementation on baseline models, and we experience performance degradation compared to the original implementation due to difference in a training scheme.
In our model, when $\tau=0.6$ is used, the accuracy difference is $0.6\%$ compared to when $\tau=1$, and the reuse rate is $25\%$.
The reuse rate is lower than the rate in DAVIS datasets for the same $\tau$. 
We assume this is partially due to the fact that Youtube-VOS dataset contains faster moving objects compared to DAVIS datasets.
As shown in Fig.~\ref{fig:teaserIntro2}, fewer consecutive frames are similar to each other than DAVIS datasets.

For further accuracy gain, we changed our training schemes of Youtube-VOS experiments.
Firstly we give relaxation to the the margin from $M(1,0)$ to $M(0.5,0)$.
Secondly, we increase the number of epochs used in the training process by additional pre-training the model without proposed modules. 
Here, our modules include the reuse gate function, the \textit{template matching module}, the \textit{delta-generato}, and the \textit{refine-translator}.
\sh{We provide following two reasons to the modification we made.}
1) Youtube-VOS Train set has less similar adjacent frames than DAVIS dataset as shown on Fig. 1(a) in Sec. 1.
2) When the gate is on, the model does not use a sub-network to make $R16_{t}$ and $R8_{t}$.
Therefore, for the sub-network to get trained equally, we need training time for the model without \sh{the} reusing process.
The overall model accuracy \sh{with the proposed modification (Tab. \ref{tab:ytb_supp}) is} better than without the pre-training stage (Tab. \ref{tab:ytb}).  
Also, the accuracy of the unseen category has not changed with different values of the threshold.

\subsection{Qualitative Examples} 
We provide our qualitative examples on a single object (DAVIS 16) and multiple objects (DAVIS 17) settings.
Fig.~\ref{fig:Q_single} shows an example of frame $23-26$ of the video \textit{cows}.
We finds that the proposed model shows better robustness than the original model.
We conjecture that the template matching helps to discriminate the \jy{desired target} objects from the \jy{non-target} objects such as cow's legs from the fence.
Fig.~\ref{fig:Q_multi} shows that each object has different reuse rate depending on the movement.
The reuse rate are 0.152 for the red segmented dog, 0.439 for the green segmented dog, and 0.864 for the yellow segmented human.
Since dogs move faster and the human does not move as much as dogs.
our gate function works properly depending on the movement of each object in the video.

%% file: Texs/Conc_sh.tex
\section{Conclusion}
\label{sec:conclusion_sh}
Semi-VOS is a task where models generate a target mask for every single frame of videos given the ground truth mask of the first frame.
Previous works on semi-VOS have treated every frame with the same importance, and this incurs redundant computation when the target object is stationary or slow-moving.
In this paper, we propose a general dynamic network that skips sub-network by quantifying the movement of targets across frames.
To do this, we estimate movement by calculating the dissimilarity between consecutive video frames using a template matching module.
Then, we train the model to learn when to skip layers of the network using a reuse gate function. 
We also propose a novel gate probability loss that takes IoU into between the previous and the current ground truth masks into account.
This loss forces the model to learn when to turn the reuse gate on, based on how similar the previous and the current frame are with preserving original accuracy.
% the reuse gate on when the previous and the current frame are substantially similar and preserving original accuracy.
Our model achieves a boosted inference speed compared to previous state-of-the-art models without significant accuracy degradation on standard semi-VOS benchmark datasets: DAVIS 16, DAVIS 17, and YouTube-VOS with multiple architectures.
% We hope that this work casts a new perspective of approaching the not only VOS task but also other video rea with dynamic inference. 
We hope that this work casts a new perspective of applying dynamic inference on not only the semi-VOS task, but also on other video-level tasks.